\pdfoutput=1

\documentclass[11pt]{article}

\usepackage{acl}

\usepackage{caption}
\usepackage{subcaption}

\usepackage{times}
\usepackage{latexsym}

\usepackage[T1]{fontenc}

\usepackage[utf8]{inputenc}

\usepackage{microtype}

\usepackage{inconsolata}
\usepackage{multirow}  
\usepackage{amssymb}   
\usepackage{graphicx}  
\usepackage{amsmath}
\usepackage{makecell} 
\usepackage{booktabs} 
\usepackage{enumitem} 
\usepackage{multirow} 
\usepackage{booktabs} 
\usepackage{stfloats}

\title{DialogVCS: Robust Natural Language Understanding \\in Dialogue System Upgrade}

\author{
Zefan Cai$^{1, 2}$\footnotemark[1],
Xin Zheng$^{3,5}$\footnotemark[1],
\textbf{Tianyu Liu}$^{4}$\footnotemark[1] \footnotemark[2],
\textbf{Xu Wang}$^{4}$,
\textbf{Haoran Meng}$^{1,2}$, \\
\textbf{Jiaqi Han}$^{4}$,
\textbf{Gang Yuan}$^{4}$,
\textbf{Binghuai Lin}$^{4}$,
\textbf{Baobao Chang}$^{1,2}$\footnotemark[2]
\textbf{and}
\textbf{Yunbo Cao}$^{4}$
\\
$^1$National Key Laboratory for Multimedia Information Processing, Peking University \\ 
$^2$School of Software and Microelectronics, Peking University, China \\
$^3$Institute of Software, Chinese Academy of Sciences, China $^4$Tencent Cloud AI \\
$^5$University of Chinese Academy of Sciences, China \\
\texttt{zefncai@gmail.com}; \texttt{zhengxin2020@iscas.ac.cn}; \texttt{rogertyliu@tencent.com}; \\
}

\begin{document}
\maketitle
\begin{abstract}
In the constant updates of the product dialogue systems, we need to retrain the natural language understanding (NLU) model as new data from the real users would be merged into the existent data accumulated in the last updates. 
Within the newly added data, new intents would emerge and might have semantic entanglement with the existing intents, e.g. new intents that are semantically too specific or generic are actually subset or superset of some existing intents in the semantic space, thus impairing the robustness of the NLU model.
As the first attempt to solve this problem, we setup a new benchmark consisting of 4 Dialogue Version Control dataSets (DialogVCS). We formulate the intent detection with imperfect data in the system update as a multi-label classification task with positive but unlabeled intents, which asks the models to recognize all the proper intents, including the ones with semantic entanglement, in the inference.
We also propose comprehensive baseline models and conduct in-depth analyses for the benchmark, showing that the semantically entangled intents can be effectively recognized with an automatic workflow\footnote{We will open source our code and data after the anonymity period.}.
\end{abstract}

\renewcommand{\thefootnote}{\fnsymbol{footnote}}
\footnotetext[1]{Equal contribution.}
\footnotetext[2]{Corresponding authors.}
\renewcommand{\thefootnote}{\arabic{footnote}}

\section{Introduction}
With the rapid growth of the business market for the task-oriented chatbots, the service providers would constantly upgrade their dialogue systems in order to be adaptable to the changing user requirements. Within the system update, the workflow of updating the existing natural language understanding (NLU) model is to collect a new training corpus by accumulating emerging data and then merging them into the existent training data in the last iteration, followed by model retrained with the updated corpus.
Throughout the model update, new intents would emerge as more and more real-world user queries arrive. 

\begin{figure}[t]
    \centering
    \includegraphics[width=0.9\linewidth]{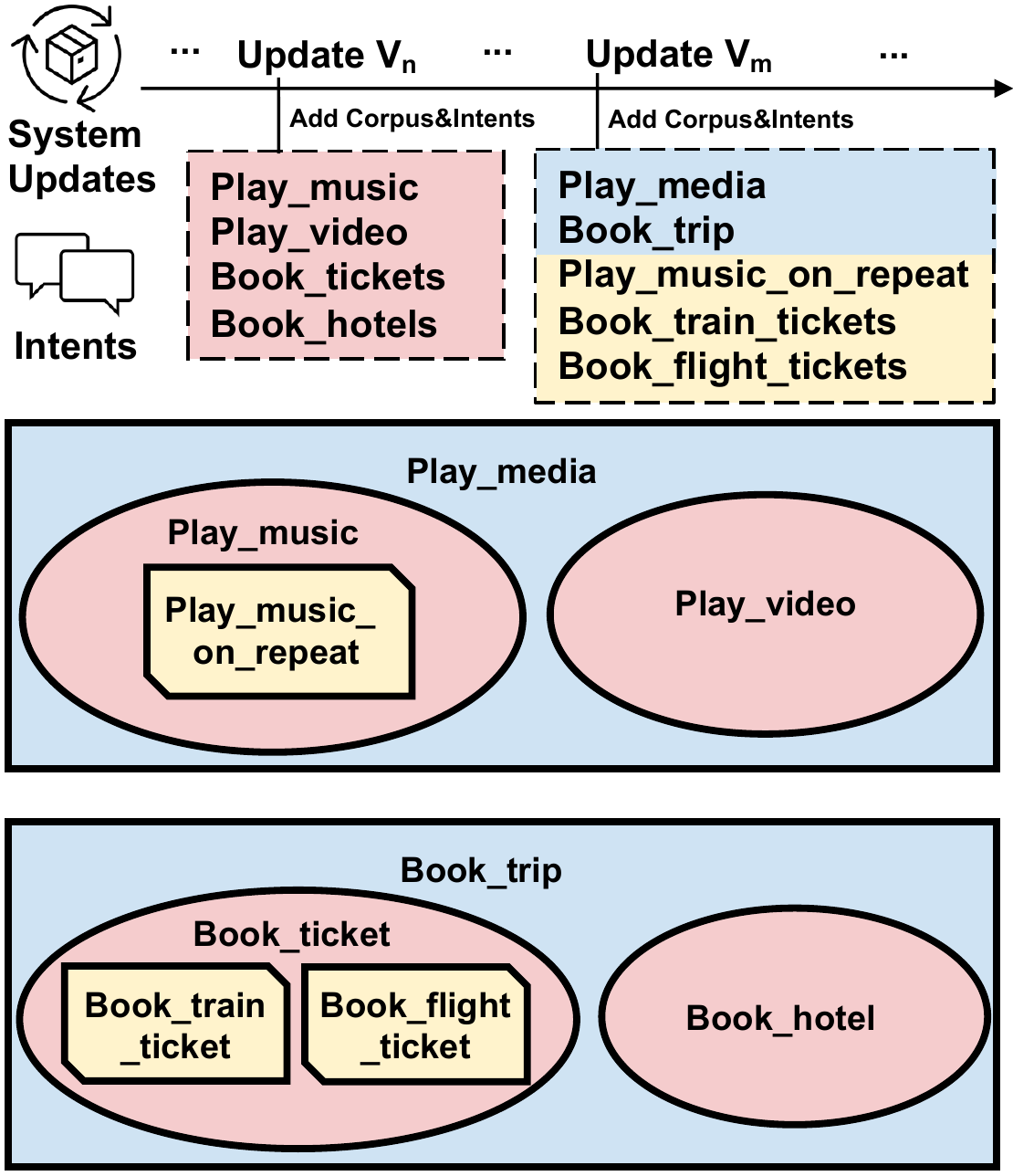}    
    \caption{
    A motivating example for DialogVCS. In the $m$-th system update, the intents colored in pink are the existing labels while the intents colored in blue and yellow are the emerging ones. The emerging intents might be overlapped, e.g. being excessively specific (yellow) or generic (blue), with the existing ones in the semantic space.
    }
    \label{fig:intro}
\end{figure}

The prior research on NLU focused on the utterance understanding with a well-defined intent\footnote{``intent'' refers to the underlying goal or purpose of a user's request or query in a dialogue. This is a commonly used concept in task-oriented dialogue datasets including MultiWOZ, CrossWOZ, SNIPS, and ATIS.} ontology, with the assumption that the entire intents are semantically separable and organized in the proper granularity\footnote{A well-designed NLU ontology should adequately split the entire user semantic space into the non-overlapping intents with appropriate granularity, i.e. each intent should not be excessively generic or specific in terms of semantics. }. However, the emerging intents from NLU model update might be incompatible with the existent intent ontology and thus violate the assumption regarding to the properties of being semantically non-overlapping and maintaining well-designed granularity, e.g. the emerging intents `\emph{play\_music\_on\_repeat}' and `\emph{play\_media}' are semantically too specific or generic with respect to the existing intent `\emph{play\_music}'.
We categorize the semantic overlapping problem between the emerging and the existing intents among the system upgrade into two categories, namely \emph{version conflict} and \emph{merge friction}, in which the version conflicts signify the emerging intents are too semantically specific and thus covered by the existing intents while the merge frictions are just the opposite.
We argue that the semantic overlapping problem between emerging and existing intents occurs frequently in the dialogue system updates as the careful human modification for each emerging intent would be prohibitive due to the limited labor budgets and the imminent product delivery deadlines. The defective data would even propagate and accumulate through consecutive upgrades, and thus largely impair the robustness of the NLU models.

We formulate the problem as a multi-label classification task with positive but unlabeled intents\footnote{We focus on the intent detection rather than slot filling, as empirically we observe over 95\% of bad cases associated with NLU model update are at the intent level in a commercial dialogue platform with a considerable market share.}. 
As the first step towards solving this problem, we setup a benchmark consisting of 4 dialogue version control datasets (DialogVCS) to simulate the semantically overlapped intents. We employ a fully automatic workflow to create the ATIS-VCS, SNIPS-VCS, MultiWOZ-VCS, CrossWOZ-VCS datasets from 4 canonical NLU datasets, including ATIS \citep{hemphill-etal-1990-atis}, SNIPS \citep{DBLP:journals/corr/abs-1805-10190}, MultiWOZ \citep{zang-etal-2020-MultiWOZ} and CrossWOZ \citep{zhu-etal-2020-CrossWOZ}, by splitting the original intents according to the pivot entities or intentions. 
The most critical challenge of DialogVCS is the discrepancy between training and inference, i.e. for each training instance, only one intent is provided as the target label\footnote{Note that we assume all the labeled intents in the training instances are factually correct, i.e. no dataset noise (false annotations) occurs. }, while in the testing phase, the models are expected to output all the ground-truth labels.
Thus we setup multiple baselines concerning with positive but unlabeled (PU) learning for the proposed benchmark and find that the baseline models are capable of detecting semantically overlapped intents in an automatic fashion.


We summarize our contributions below:
\textbf{1)} We model the version conflicts and merge frictions of NLU models in the industrial dialogue system update as a multi-label classification task with positive but unlabeled intents, making it accessible to the research community. 
\textbf{2)} We propose 4 dialogue version control datasets by simulating the semantic overlapping problem on the ATIS, SNIPS, MultiWOZ, and CrossWOZ datasets.
\textbf{3)} We setup various baselines for the proposed benchmark and show that the semantically overlapping intents can be effectively detected with an automatic workflow.

\section{Task Overview}
\paragraph{Background on system updates} In the product conversational AI platforms with NLU functionalities
\cite{ram2018conversational,hoy2018alexa,meng-etal-2022-dialogusr,zheng2022dialogqae,liang2022smartsales} based on cloud computing, service providers would offer accessible ways, i.e. easy-to-use user interfaces, low-code application programming interfaces (APIs), for users (programmers or operators) to customize their task-oriented dialogue systems.
As one of the core components in the task-oriented chatbots, the dialogue platform would provide common query understanding skills, such as weather and traffic inquiry, music and video playing, and food delivery, as the default native skills to ramp up the initial product delivery. The native skills would be updated periodically as more and more customer data comes from real-world users.
After deploying the very first version of their chatbots with selected native skills, the users would constantly add new functionalities or modify existing ones following the continuous integration/delivery (CI/CD) routines \cite{duvall2007continuous,shahin2017continuous}.
Except for the native skills, users would also customize user skills by adding their own training corpus\footnote{Most AI platforms would help the users reduce the labor cost of data annotation with automatic data augmentation, few-shot learning capability, etc.} to the platform.
In a nutshell, the natural language understanding (NLU) module of the task-oriented chatbots might be updated due to \emph{the upgrades of the native skills} or \emph{the adaptations to the customized user skills}.

\paragraph{Formulations} To better signify the two aforementioned challenges, suppose at first we have two intents $i_1$ and $i_2$, the version conflict would occur when the new intents $i_1^{v1}$, $i_1^{v2}$ emerges where the superscripts ${v1}$ and ${v2}$ imply that $i_1^{v1}$ and $i_1^{v2}$ are different labels with respect to $i_1$ but semantically identical; the merge friction would occur as the new intent $i_1\&{i_2}$ appears where the ampersand emphasizes the new intent is different but semantically affiliated to $i_1$ and $i_2$. Note that $i_1$, $i_1^{v1}$ and $i_1\&{i_2}$ are just the notations of the given intents rather than the real intent names, which means we can not know the relations among these intents a prior.






\section{Dataset Collection}
\subsection{Raw Data Collection}
We collect data from two single-turn dialog datasets ATIS \citep{hemphill-etal-1990-atis} and SNIPS \citep{DBLP:journals/corr/abs-1805-10190}, and two multi-turn dialog datasets MultiWOZ 2.1 \citep{zang-etal-2020-MultiWOZ} and CrossWOZ \citep{zhu-etal-2020-CrossWOZ}. ATIS is a classic dataset on the flight inquiry, while SNIPS was collected from the real-world voice assistant and covers broader domains. MultiWoZ is a task-oriented dataset with seven domains: taxi, restaurant, hotel, attraction, train, police, and hospital, but the last two domains are not in the validation or test set, so we drop them following the prior work \citep{lee-etal-2019-sumbt, kim-etal-2020-efficient, DBLP:journals/corr/abs-2111-02574}. CrossWOZ is a Chinese task-oriented dataset with the same domain setting as MultiWOZ's validation/test set: taxi, restaurant, hotel, attraction, and train. For these WOZ datasets, we treat each utterance as an instance, rather than the whole dialog. The statistics of the datasets is shown in Table \ref{table:statistics}.

\subsection{Version Conflict}
We simulate the version conflict by sampling. Given an instance $Ins$ with the original label $l=i_1$ and versions set $V=\{ v_1, v_2, ..., v_k\}$, we uniformly sample the version $v$ from $V$, and reset the label of the instance as $l'=i_1^v$. In the real-world applications, a specific intent might have multiple versions, but to control the difficulty of the dataset, here we assume the maximum number of versions is 2, i.e. $k=2$. At testing time, the model shall predict both versions of the label $i_1^{v_1}$ and $i_1^{v_2}$.

\subsection{Merge Friction}
For merge friction, the label splitting strategies on composite intents are different regarding to single-intent and multi-intent datasets. 

\begin{figure}
     \centering
     \begin{subfigure}[b]{0.45\textwidth}
         \centering
         \includegraphics[width=\textwidth]{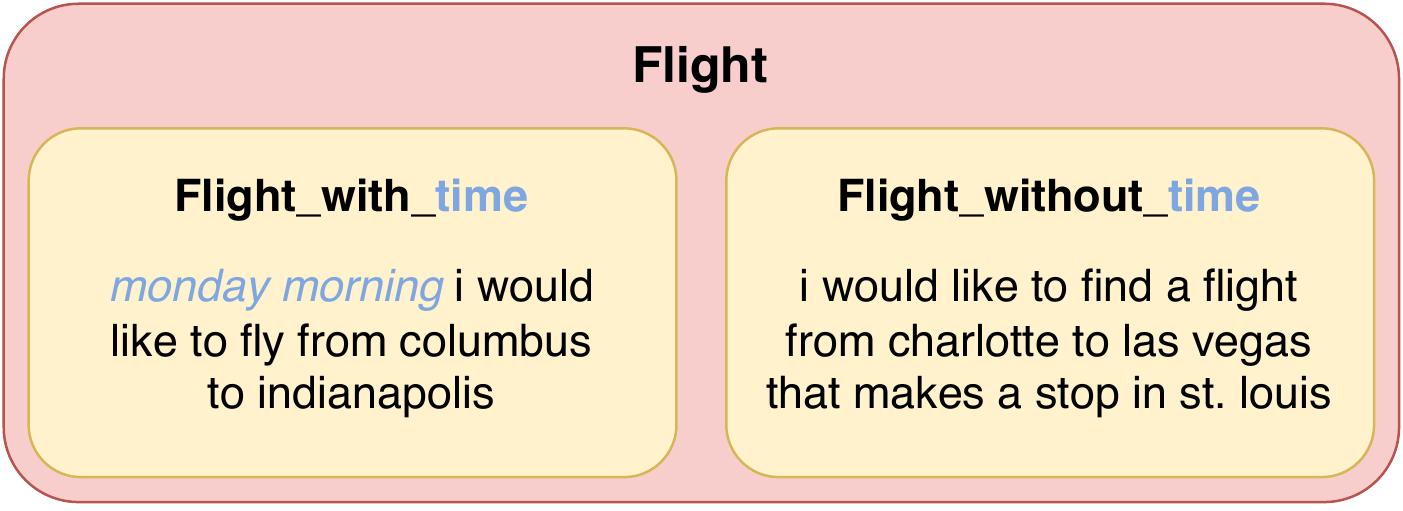}
         \caption{Split by entity}
         \label{fig:split atis}
     \end{subfigure}
     \begin{subfigure}[b]{0.45\textwidth}
         \centering
         \includegraphics[width=\textwidth]{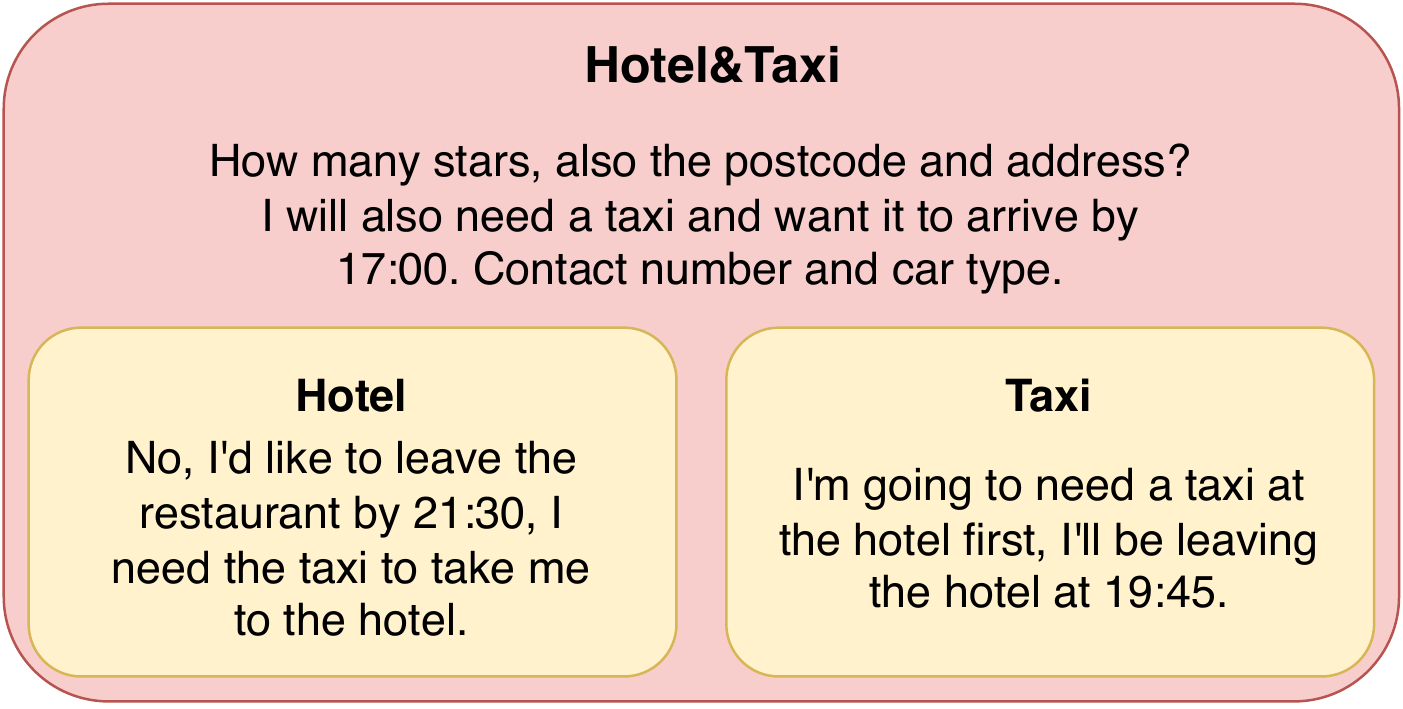}
         \caption{Split by intention}
         \label{fig:split multiwoz}
     \end{subfigure}
        \caption{The examples of intent splitting while simulating the merge friction issue. (\ref{fig:split atis}) For single-intent datasets, i.e. ATIS and SNIPS, we split the intent ``$Fight$'' into two sub-intents ``$Flight\_with\_time$'' and ``$Flight\_without\_time$'' by the pivot entity ``$time$''. (\ref{fig:split multiwoz}) For multi-intent datasets, i.e. MultiWOZ and CrossWOZ, we split the composite intent ``$Hotel\&Taxi$'' into two atomic intents ``$Hotel$'' and ``$Taxi$''.}
        \label{fig:split examples}
\end{figure}

\paragraph{Split Single Intent} For ATIS and SNIPS, where each instance is annotated with one single intent $i$ and several related entities $E$ or slots, we could split the single intent $i$ into two separate sub-intent $i_1 = i_{with\_entity\_j}$ and  $i_2 = i_{without\_entity\_j}$, the classification rule of which is whether this instance contains the $entity\_j$ or not, and the original intent $i$ becomes compositional $i_1 \& i_2$. For example, as shown in figure \ref{fig:split atis}, given an utterance ``i would like to find a flight from charlotte to las vegas that makes a stop in st. louis'' with the intent $Flight$, since it does not contain any time entity, the sub-intent shall be $Flight\_without\_time$; on the other hand, given an utterance ``monday morning i would like to fly from columbus to indianapolis'' with the same intent, since it contains time entity ``monday morning'', the sub-intent shall be $Flight\_with\_time$. For training data, we randomly relabel the instance by sub-intent $i_1, i_2$ or full-intent $i_1\&i_2$. While testing, the model shall predict both the fine-grain and coarse-grain labels. The split intents and their dividing entity for ATIS-VSC and SNIPS-VSC is shown in Table \ref{table:split_intent}.

\paragraph{Split Multi Intent} 
Unlike the previous situation, for MultiWOZ and CrossWOZ each instance might contain multiple intents, which makes splitting intent easier. We reconsider the deduplicated multi-intents as a new compositional label $i_1\&i_2$, and naturally its atomic labels are $i_1$ and $i_2$. An example is shown in Figure \ref{fig:split multiwoz}, each of the three instances could be labeled as any of the three labels, whether compositional label $Hotel\&Taxi$, or atomic labels $Hotel$ and $Taxi$. For training data, we randomly relabel the instance by one of the atomic intents $i_1$ and $i_2$, or the compositional intent $i_1\&i_2$. While testing, the model should predict all the ground-truth labels. The split intent is presented in Table \ref{table:split_multiwoz_crosswoz} in Appendix.




\section{Methods}
\label{section:methods}

We highlight the technical challenges of DialogVCS: 
\textbf{1)} The discrepancy between training and testing due to the positive but unlabeled (PU) setting; 
\textbf{2)} The risk of pivoting the model training with false negative labels;
\textbf{3)} The extreme 0-1 class imbalance of multi-label classification.
We propose multiple baselines towards these challenges.

\subsection{Basic Classifier}
\label{subsection:basic_classifier}

Considering the proposed task as a multi-label classification task, we apply a linear classification at the head of the output of pre-trained language model (PLM).
we use a PLM to get the representations for every token $x$ in sentence: $[\mathbf{h}_1,\mathbf{h}_2,...,\mathbf{h}_n] = PLM([\mathbf{x}_1,\mathbf{x}_2,...,\mathbf{x}_n])$
where $h_i$ is the representation for token $x_i$. 
Then, we use linear transformation and $Sigmoid$ activation function at the output representation of $[CLS]$ to get output distribution for intents: $\mathbf{y} = Sigmoid(W\textbf{h}_1)$,
where $W$ is trainable parameter. In practice, we use threshold $0.5$ for the output of $Sigmoid$ to determine the final binary output for each intent.

\subsection{Method against False Negative Labels}
\label{subsection:method_against_incorrect_negative_label}

In order to alleviate the negative effect of false negative labels, we propose Negative Sample method to reduce the negative effects of the inaccurate negative samples. For each sample $s$ in training set $D_{train}$, instead of directly using the labels given by dataset, we construct new labels by using the positive label and randomly sample $\theta * |L|$ negative samples, where $theta$ is a proportion and $|L|$ is the number of labels of the dataset. We use the model output as the labels other than the positive label and the sampled negative labels, meaning that we do not optimize all labels other than positive and negative labels. And then we use BCE Loss~\citep{creswell2017denoising} for optimization.

\subsection{Method for Imbalanced Binary Classification}

If we consider the proposed task as intent binary classification, the distribution of positive and negative sample for each class is extremely imbalanced. Targeting at the unbalance of positive and negative sample for each intent, we propose a method based on Focal Loss with label smoothing, which puts more emphasis on positive samples. Specifically, we add a label something on the original target $l$:

\begin{equation}
l^{LS} = l(1-\beta) + \frac{\beta}{|L|}
\label{equation:label_smoothing}
\end{equation}

where $|L|$ denotes the number of intent classes. $\mathbf{\beta}$ is the label smoothing parameter. $\mathbf{\beta / K}$ is the soft label, which represents the number of intent labels. $l$ is a vector where the positive labels equal to 1 and negative labels equal to 0 and $p^{LS}$ is the modified targets, which represents a list of ground truth labels.

 We introduce Focal Loss~\citep{lin2017focal} to alleviate the above problems. For notational convenience, we define $\mathbf{p}_t$ as below:

\begin{equation}
    \mathbf{p}_t=\begin{cases} p &\text{if $y = 1$}\\ 1 - p &\text{otherwise,}\end{cases}
\end{equation}

To address class imbalance, we introduce a weighting factor $\mathbf{\alpha}_t \in [0, 1]$ for class  $1$ and $1 - \alpha$ for class $-1$. As the extreme class imbalance encountered during training of classifier overwhelms the cross entropy loss and major negative samples, the easily classified negative samples comprise the majority of the loss and dominate the gradient. As $\mathbf{\alpha}$ balances the importance of positive and negative samples, we add another factor $(1 - \mathbf{p}_t)^\gamma$ to differentiate between easy and hard samples and focus training on hard negatives:

\begin{equation}
    FL(\mathbf{p}_t) = - \mathbf{\alpha}_t (1 - \mathbf{p}_t)^\gamma \log ( \mathbf{p}_t).
    \label{equation:focal_loss}
\end{equation}

where $\mathbf{\alpha}$ and $\mathbf{\gamma}$ are hyper parameters. Considering the proposed task as binary classification, there are 2 hyper-parameters $\mathbf{\alpha}_{\text{pos}}$ and $\mathbf{\alpha}_{\text{neg}}$ for $\mathbf{\alpha}_t$

\subsection{Method for Imbalanced Multi-Label Label Classification}
\label{subsection:method_for_imbalanced_multi_label_label_classification}

Another method that we are interested to explore is to apply Cross Entropy Loss into multi-label classification instead of modeling the proposed task as binary classification. Cross Entropy Loss maximize the difference between the score of target class and the score of other classes:

\begin{equation}
    L_{CE} = \log \left(1+\sum_{i=1, i \neq t}^{n} e^{s_{i}-s_{t}}\right)
\end{equation}

where $[\mathbf{x}_1, \cdots, \mathbf{s}_{t-1}, \mathbf{s}_{t+1}, \cdots, \mathbf{s}_n]$ is the output score of non-target classes and $\mathbf{s}_t$ is the output score of target class.
As an extension to apply CE Loss at multi-label classification, we still want to maximize the difference between the score of target classes and the score of other classes, so we propose a multi-label CE Loss:

\begin{equation}
\begin{aligned}
    L_{mlCE} &= \log \left(1+\sum_{\mathrm{i} \in \Omega_{\text {neg }}, \mathrm{j} \in \Omega_{\text{pos}}} e^{\mathrm{s_i}-\mathrm{s_j}}\right) \\
             &= \log \left(1+\sum_{i \in \Omega_{\text{neg}}} e^{s_i} \sum_{j \in \Omega_{\text{pos}}} e^{-s_j}\right)
\end{aligned}
\label{equation:multi-label-cross-entropy}
\end{equation}

where $\Omega_{\text{neg}}$ denotes negative classes and $\Omega_{\text{pos}}$ denotes positive classes. The optimized goal of $L_{mlCE}$ is to make $s_i \textless s_j$.

In our proposed task, the number of output classes is unfixed, so we need a threshold to determine which class to be positive. We introduce an additional threshold score $\mathbf{x}_0$ and optimize to make $\mathbf{s}_j \textgreater \mathbf{s}_0$ and $\mathbf{s}_i \textless \mathbf{s}_0$ into Equation~\ref{equation:multi-label-cross-entropy}:

\begin{equation}
\begin{aligned}
    L_{mlCE}  = &\log \left(e^{s 0}+\sum_{i \in \Omega_{n e g}} e^{s_{i}}\right) \\ + &\log \left(e^{-s_{0}}+\sum_{j \in \Omega_{p o s}} e^{-s_{j}}\right)
\end{aligned}
\label{equation:multi-label-cross-entropy-threshold}
\end{equation}

Equation~\ref{equation:multi-label-cross-entropy-threshold} is the extension of Softmax and Cross Entropy to multi-label classification task. 
Instead of turning multi-label classification into multiple binary problem, it transforms it to a two-by-two minimization of scores of target classes with non-target classes, leading to alleviation of class unbalance. As we use threshold $0.5$ for the output of $Sigmoid$ to determine the final binary output for each intent, we set $s_0$ to be $0$.

\subsection{Method of In-Context-Learning}
\label{subsection:method_of_in_context_learning}

Large Language Models (LLMs) ~\citep{sanh2021multitask,ouyang2022training, zhang2022opt} have demonstrated impressive few-shot generalization abilities. We are also interested in investigating generation-based methods and incorporating label semantics as inputs for generative models. For each dataset, we provide one data sample for each label. We also provide a task description and all the available label options and query the generative model to output one or more labels that match the input.

\section{Experiments}
\label{section:experiments}



\begin{table*}[!htp]
\centering
\small
\begin{tabular}{llllll|lll}
\toprule
\multirow{2}{*}{\textbf{Dataset}} & \multicolumn{5}{c|}{\textbf{Intent Statistics}} & \multicolumn{3}{c}{\textbf{Dataset Count}} \\
 & \textbf{VC-N} & \textbf{VC-R(\%)} & \textbf{MF-N} & \textbf{MF-R(\%)} & \textbf{Total} & \textbf{Train} & \textbf{Valid} & \textbf{Test} \\
\midrule
ATIS-VCS & 50 & 75.8 & 10 & 15.2 & 66 & 4455 & 496 & 876 \\
SNIPS-VCS & 24 & 77.4 & 6 & 19.4 & 31 & 13084 & 700 & 700 \\
MultiWOZ-VCS & 14 & 63.6 & 8 & 36.4 & 22 & 42342 & 4229 & 4238 \\
CrossWOZ-VCS & 10 & 58.8 & 7 & 41.2 & 17 & 55189 & 7325 & 7305 \\
\bottomrule
\end{tabular}
\caption{
The statistics of the proposed datasets. We list the label number of the intents which involves the version conflict (VC-N) or the merge friction (MF-N) issues, the correlated ratio of concerning training instances in the training set (VC-R and MF-R), as well as the dataset split for  training, validation and testing.
}
\label{table:statistics}
\end{table*}

\subsection{Datasets and Evaluation Metrics}
\label{subsection:evaluation_metrics}
We show the dataset statistics in Table \ref{table:statistics}.
To compare the baseline models, we adopt the standard precision(P), recall(R), F1-score(F1) for evaluation. The above metrics consider the task as a binary classification task for all intents, ignoring the multi-label classification nature of the task. So we present the exact match ratio (EM) metrics for further evaluation. All the above metrics are under the setting that a label is predicted as positive if its estimated probability is greater than $0.5$~\citep{zhu2017learning}. Among these metrics, F1 and EM are the most representative metrics.

\begin{small}
\begin{equation}
   \begin{split}
         \mathrm{P}&=\frac{\sum_{i}N_{i}^{c}}{\sum_{i}N_{i}^{p}},\quad\\
         \mathrm{R}&=\frac{\sum_{i}N_{i}^{c}}{\sum_{i}N_{i}^{g}},\quad\\
   \end{split}
      \begin{split}
         \mathrm{F}1&=\frac{2 \times \mathrm{P} \times \mathrm{R}}{\mathrm{P}+\mathrm{R}},\quad\\
         \mathrm{EM} &= \frac{1}{m} \sum_{j=1}^{m} I\left({p_j==l_j}\right)\quad
   \end{split}
   \label{eqn:metric}
\end{equation}
\end{small}%

where $N_{i}^{c}$ is the number of intents that are correctly predicted to be true for the $i$-th label, $N_{i}^{p}$ is the number of intents predicted to be true for the $i$-th label, $N_{i}^{g}$ is the number of ground truth intents for the $i$-th label, $m$ is the number of instances in test dataset $D_{test}$, $p_j$ is the model output of all intent labels for sample $s_j$, $l_j$ is the ground truth intent labels for sample $s_j$ and $I()$ is an indicator function, which will output 1 when the distribution of $p_j$ is equivalent to $l_j$.

\subsection{Experiment Settings}
\label{section:experiment_settings}

For a fair comparison, we use BERT-base-uncased~\citep{devlin2018bert} as the text encoder for all methods. 
We introduce a naive baseline by applying a basic multi-label classifier (Section~\ref{subsection:basic_classifier}) to the proposed task.
Another baseline we introduce is to train the classifier exposure to all ground-truth labels, which indicate the upper bound of other models as all other models is trained with partially positive labels.

We implement all the experiment with Hugginface Transformers \cite{wolf-etal-2020-transformers}. We specify the model\_ids we used in the model repository in Table \ref{tab:appendix_model_map}. All the hyper parameters used in our proposed methods are presented in Table~\ref{table:appendix_hyper_parameters}.

\subsection{Experiment Results}
\label{section:main_results}

\begin{table*}
\centering
\small
\setlength\tabcolsep{2.2pt}
    \centering
\begin{tabular}{l|cccc|cccc|cccc|cccc}
\toprule
\multirow{2}{*}{\textbf{Method}}  & 
\multicolumn{4}{c|}{\textbf{ATIS-VCS}}  & 
\multicolumn{4}{c|}{\textbf{SNIPS-VCS}} &
\multicolumn{4}{c|}{\textbf{CrossWOZ-VCS}} &
\multicolumn{4}{c}{\textbf{MultiWOZ-VCS}} \\
  & P & R & F1 & EM & P & R & F1 & EM & P & R & F1 & EM & P & R & F1 & EM \\
\midrule
Basic classfier & 66.67 & 0.01 & 0.15 & 0.00 & \textbf{99.99} & 5.26 & 10.00 & 14.29 & \textbf{98.06} & 23.83 & 38.35 & 3.94 & 91.78 & 37.93 & 53.67 & 6.75 \\
\midrule
Neg. Sample & 87.4 & 86.87 & 87.14 & 76.37 & 94.30 & 93.16 & 93.73 & 85.14 & 97.97 & 49.24 & 65.54 & 42.97 & 86.19 & \textbf{87.06} & 86.62 & 82.79 \\
LS Focal loss & 84.17 & \textbf{88.81} & 86.43 & 77.05 & 95.85 & \textbf{95.95} & \textbf{95.90} & \textbf{92.86} & 97.00 & \textbf{88.37} & \textbf{92.48} & \textbf{80.34} & 88.62 & 86.45 & \textbf{87.52} & \textbf{85.85} \\
Multi-label CE & \textbf{91.77} & 85.73 & \textbf{88.65} & \textbf{79.91} & 94.40 & 80.74 & 87.04 & 65.14 & \textbf{98.06} & 28.86 & 44.60 & 14.47 & \textbf{94.00} & 81.14 & 87.10 & 80.46 \\
\midrule
ChatGPT-ICL & 49.84 & 52.33 & 51.06 & 0.03 & 82.86 & 0.58 & 0.6824 & 31.79 & 11.37 & 16.75 & 0.01 & 42.97 & 60.92 & 51.46 & 55.79 & 1.00 \\
\midrule
Upper Bound & \textbf{98.07} & 86.80 & \textbf{92.09} & \textbf{83.22} & 96.73 & \textbf{96.42} & \textbf{96.57} & \textbf{95.86} & 96.90 & \textbf{96.95} & \textbf{96.92} & \textbf{93.49}  & 89.33 & \textbf{87.34} & \textbf{88.32} & \textbf{86.71}\\
\bottomrule
\end{tabular}
    \caption{ 
Model performance on the DialogVCS. We use BERT-base as the backbone text encoder for all the baselines. The `Basic Classifier' and `Upper Bound' methods signify the `know nothing a priori' (no inductive bias of positive but unlabeled (PU) learning in the training) and `know everything a priori' (exposure to all ground-truth labels in the training) settings, while other methods aim to recognize unlabeled intents in the regime of PU learning.
For each setting except ChatGPT-ICL, we report the median scores among 5 runs. 
}
    \label{table:main-results}
\end{table*}

\paragraph{Main Results} As shown in Table \ref{table:main-results}, due to the discrepancy between the label distribution in the training and testing, fine-tuning the classifier by the naive method of `Basic Classifier' as Sec.~\ref{subsection:basic_classifier} to DialogVCS with the naive BCE Loss yields low performance, indicating the challenges of DialogVCS.
The proposed baselines significantly alleviate the negative effect of inaccurate negative labels. Among the three methods, Multi-Label Focal Loss as Sec.~\ref{subsection:method_for_imbalanced_multi_label_label_classification} generally outperforms other methods to be a robust method for partial positive labels. 

For new intents that have no semantic overlapping with the original intents, we train them directly as new samples without considering version conflicts or merge frictions. Since these new intents do not overlap semantically with the original intents, they can be added to the training data without any issues.

We experimented with in-context learning of GPT-3.5 ~\footnote{\url{https://openai.com/blog/chatgpt}}. We provide one sample for each intent in the demonstration to form many examples (i.e., 66 intents for ATIS-VCS, 31 intents for SNIPS-VCS, 22 intents for CrossWOZ-VCS, and 17 intents for MultiWOZ-VCS). We add the requirement of completing the multi-label classification task and provide all options in the prompt. Then we determine the intent of the model output by matching the options provided in the prompt with the generated text output. Following ~\citet{ye2023comprehensive, qin2023chatgpt}, we randomly sample 100 instances in the test set for the test. The performance of GPT-3.5 on in-context learning ~\citep{kojima2022large} under few-shot settings is satisfactory enough, which further demonstrates the challenging nature of the proposed benchmark.

\paragraph{Analysis on how to address the problem of intentional overlap in new and old data}
\label{paragraph:analysis_on_how}

The benchmark can be seen as a unique adversarial dataset. It contains both test and training data, allowing for the analysis of model performance and trends under different levels of inconsistency control. This approach helps reveal the robustness of the model. As demonstrated in Table 5, the naive basic classifier baseline experiences a significant drop in performance as the data becomes more inconsistent. However, a robust model should ideally not exhibit such a rapid decline in accuracy. Instead, it should generally maintain accuracy, or even approach the performance upper bound. This benchmark aims to reveal these characteristics in the tested models, contributing to the development of more robust NLU models for industrial dialogue systems.
In addition, we make contributions to the method to address this problem. Our motivation for designing the method is to model the problem as a PU learning problem of multi-label classification. Next, we want the model to be able to identify semantically overlapped intents, so we apply three methods: Negative Sampling, Label-Smoothing Focal Loss, and Multi-Label Cross-Entropy.

%

\paragraph{Model Scale Up}

Table \ref{table:scale_up} shows the model performance on DialogVCS with different size of text encoder. 
We use Label-Smoothing Focal Loss method due to its high performance in Table \ref{table:main-results}.
Results show that scaling up generally benefits the model performance. Transferring from BERT-Small to BERT-Base brings up to 9 points growth in the F1 score, and transferring from BERT-Base to BERT-Large brings up to 5 points growth in the F1 score. However, the performance of CrossWOZ-VCS dataset does not follow this trend, which might be caused by the insufficient training of large-size Chinese BERT models.

\begin{table}[ht!]
\centering
\small
\setlength\tabcolsep{2.2pt}
\begin{subtable}[]{0.5\textwidth}
\centering
\begin{tabular}{l|c|c|c|c}
\toprule

\multirow{1}{*}{\textbf{Size}}  & 
\multicolumn{1}{c|}{\textbf{ATIS-VCS}}  & 
\multicolumn{1}{c|}{\textbf{SNIPS-VCS}} &
\multicolumn{1}{c|}{\textbf{Cro-VCS}} &
\multicolumn{1}{c}{\textbf{Mul-VCS}} \\
\midrule
Small & 77.78 & 90.68 & \textbf{95.60} & 86.26 \\
Base & 86.43 & 95.90 & 92.48 & \textbf{87.52} \\
Large & \textbf{91.57} & \textbf{97.45} & 87.66 & 87.34 \\
\bottomrule
\end{tabular}
\caption{Exploration on Model Scale}
\label{table:scale_up}
\end{subtable}

\begin{subtable}[]{0.5\textwidth}
\centering
\begin{tabular}{l|c|c|c|c}
\toprule
\multirow{1}{*}{\textbf{Model}}  & 
\multicolumn{1}{c|}{\textbf{ATI-VCS}}  & 
\multicolumn{1}{c|}{\textbf{SNI-VCS}} &
\multicolumn{1}{c|}{\textbf{Cro-VCS}} &
\multicolumn{1}{c}{\textbf{Mul-VCS}} \\
\midrule
BERT & 86.43 & 95.90 & 92.48 & \textbf{87.52} \\
RoBERTa & 91.03 & \textbf{96.34} & 92.41 & 86.62 \\
AlBERT & 84.64 & 88.45 & 84.58 & 85.92 \\
DeBERTa & \textbf{91.56} & 90.89 & \textbf{95.93} & 86.61 \\
\bottomrule
\end{tabular}
\caption{Exploration on Model Structures}
\label{table:model_structure}
\end{subtable}

\begin{subtable}[]{0.5\textwidth}
\centering
\begin{tabular}{l|c|c|c|c}
\toprule
\multirow{1}{*}{\textbf{LSR}}  & 
\multicolumn{1}{c|}{\textbf{ATIS-VCS}}  & 
\multicolumn{1}{c|}{\textbf{SNIPS-VCS}} &
\multicolumn{1}{c|}{\textbf{Cross-VCS}} &
\multicolumn{1}{c}{\textbf{Multi-VCS}} \\
\midrule
0.1 & 77.67 & \textbf{95.90} & \textbf{92.48} & 86.86  \\
0.2 & \textbf{86.43} & 95.05 & 80.89 & 87.02  \\
0.4 & 85.13 & 88.58 & 80.59 & \textbf{87.52}  \\
\bottomrule
\end{tabular}
    \caption{ 
Exploration on Label Smoothing Rates
}
\label{table:label_smothing}
\end{subtable}

\caption{
The F1 scores of the the Label-Smoothing Focal Loss method with different model size (\ref{table:scale_up}), different structures of the encoder (\ref{table:model_structure}), and different label smoothing rates (LSR) (\ref{table:label_smothing}). The full tables are provided in Table \ref{table:appendix_scale_up}, Table \ref{table:appendix_model_structure}, and Table \ref{table:appendix_label_smothing}.
}
\label{table:many_table}
\end{table}

\paragraph{Model Structure}

We are also interested in whether the selection of model structure for text encoder is important for the task performance. Table \ref{table:model_structure} shows the model performance with different model structure for text encoder. We experiment four model structures of the text encoder, including BERT-Base, RoBERTa-Base \cite{liu2019roberta}, AlBERT-Base \cite{lan2019albert} and DeBERTa-Base \cite{he2020deberta}. 
Results show that RoBERTa-Base and DeBERTa-Base generally outperform other model structures.

\paragraph{Label Smoothing Rate for Focal Loss}

Our Label-Smoothing Focal Loss method consists of a dedicated label smoothing strategy. Intuitively, as the negative samples 
 are prone to be false negative in DialogVCS, smoothing the labels in this way prevents the classifier from becoming over-confident while determining negative outputs. Table \ref{table:label_smothing} shows the model performance on DialogVCS when applying Label-Smoothing Focal Loss method with different label smoothing rates (LSR).  
The best practise for choosing label smoothing rate depends on the number of labels of the dataset, generally speaking a dataset with larger label set requires a larger label smoothing rate. As shown in table \ref{table:label_smothing}, the numbers of labels in the ATIS-VCS dataset and MultiWOZ-VCS dataset are larger than those in the SNIPS-VCS dataset and CrossWOZ dataset, thus the Label-Smoothing Focal Loss method attains better performance with a larger label smoothing rate such as $0.2$ and $0.4$, while the best choice of label smoothing rate for the SNIPS-VCS dataset and CrossWOZ-VCS dataset is $0.1$.


\paragraph{Negative Sample Number}

There is a critical hyper-parameter for the negative sampling method — the negative sample number. As illustrated in Table \ref{table:negative_sample_number}, we try to figure out the best hyper-parameter setting in terms of the negative sample number. We observe that as the negative sample number increases, the performance decreases to a large extent for the SNIPS-VCS, CrossWOZ-VCS and MultiWOZ-VCS, with an exception that 4 negative samples work the best for the ATIS-VCS dataset.


\begin{table}
\centering
\small
\setlength\tabcolsep{2.2pt}
    \centering
\begin{tabular}{l|c|c|c|c}
\toprule
\multirow{1}{*}{\textbf{NSN}}  & 
\multicolumn{1}{c|}{\textbf{ATIS-VCS}}  & 
\multicolumn{1}{c|}{\textbf{SNIPS-VCS}} &
\multicolumn{1}{c|}{\textbf{Cross-VCS}} &
\multicolumn{1}{c}{\textbf{Multi-VCS}} \\
\midrule
1  & 66.35 & \textbf{93.73} & \textbf{65.54} & \textbf{86.62}\\
2  & 79.55 & 91.67 & 58.78 & 84.57\\
4  & \textbf{87.14} & 82.37 & 52.90 & 77.59\\
8  & 84.46 & 76.99 & 48.72 & 72.60\\
\bottomrule
\end{tabular}
    \caption{ 
The F1 scores of the Negative Sampling Method under different negative sample numbers (NSN). The full table is provided in Table~\ref{table:appendix_result_difficulty_control}.
    }
    \label{table:negative_sample_number}
\end{table}

\paragraph{Difficulty Control}
\label{paragraph:difficulty_control}

We want to explore the model performances on DialogVCS with different levels of semantic entanglement. Intuitively, we can control the difficulty level by controlling the number of conflicting labels, e.g. `easy' and `hard' versions of DialogVCS. The details of creating such datasets are presented in Appendix~\ref{appendix:difficulty_control}. As shown in Table~\ref{table:result_difficulty_control}, in ATIS-VCS and SNIPS-VCS, as the number of split sub-intents decreases, the dataset becomes easier, and the performance improves. While in CrossWOZ-VCS and MultiWOZ-VCS, as the number of split atomic intents decreases, the ratio of simple intents also decreases, thus the dataset becomes harder, and the performance declines. We put more details in Table~\ref{table:appendix_result_difficulty_control}.   

\begin{table}
\small
\centering
\makeatletter\def\@captype{table}
\begin{tabular}{l|c|c}
\toprule
\multirow{1}{*}{\textbf{Difficulty}}  & 
\multicolumn{1}{c|}{\textbf{ATIS-VCS}}  & 
\multicolumn{1}{c}{\textbf{SNIPS-VCS}} \\
\midrule
Easy 1 & 96.17 & 98.50 \\
Easy 2 & 96.46 & 95.93 \\
Easy 4 & 93.15 & 96.72 \\
Normal & 86.43 & 95.90 \\
\bottomrule
\toprule
\multirow{1}{*}{\textbf{Difficulty}}  & 
\multicolumn{1}{c|}{\textbf{CrossWOZ-VCS}}  & 
\multicolumn{1}{c}{\textbf{MultiWOZ-VCS}} \\
\midrule
Hard 1 & 76.07 & 84.96 \\
Hard 2 & 80.89 & 85.41 \\
Hard 4 & 80.59 & 86.29\\
Normal & 92.48 & 87.52 \\
\bottomrule
\end{tabular}
\caption{
The F1 scores of the Label-Smoothing Focal Loss method with different levels of difficulty. We control the dataset difficulty by controlling the group numbers of label versions, i.e. $k$ in ``Easy $k$'' or ``Hard $k$'' (Appendix~\ref{appendix:difficulty_control}). 
}
\label{table:result_difficulty_control}
\end{table}

\paragraph{Correlation Between Labels}
\label{paragraoh:correction_between_labels}

\begin{figure*}[t]
    \centering
    \includegraphics[width=0.9\linewidth]{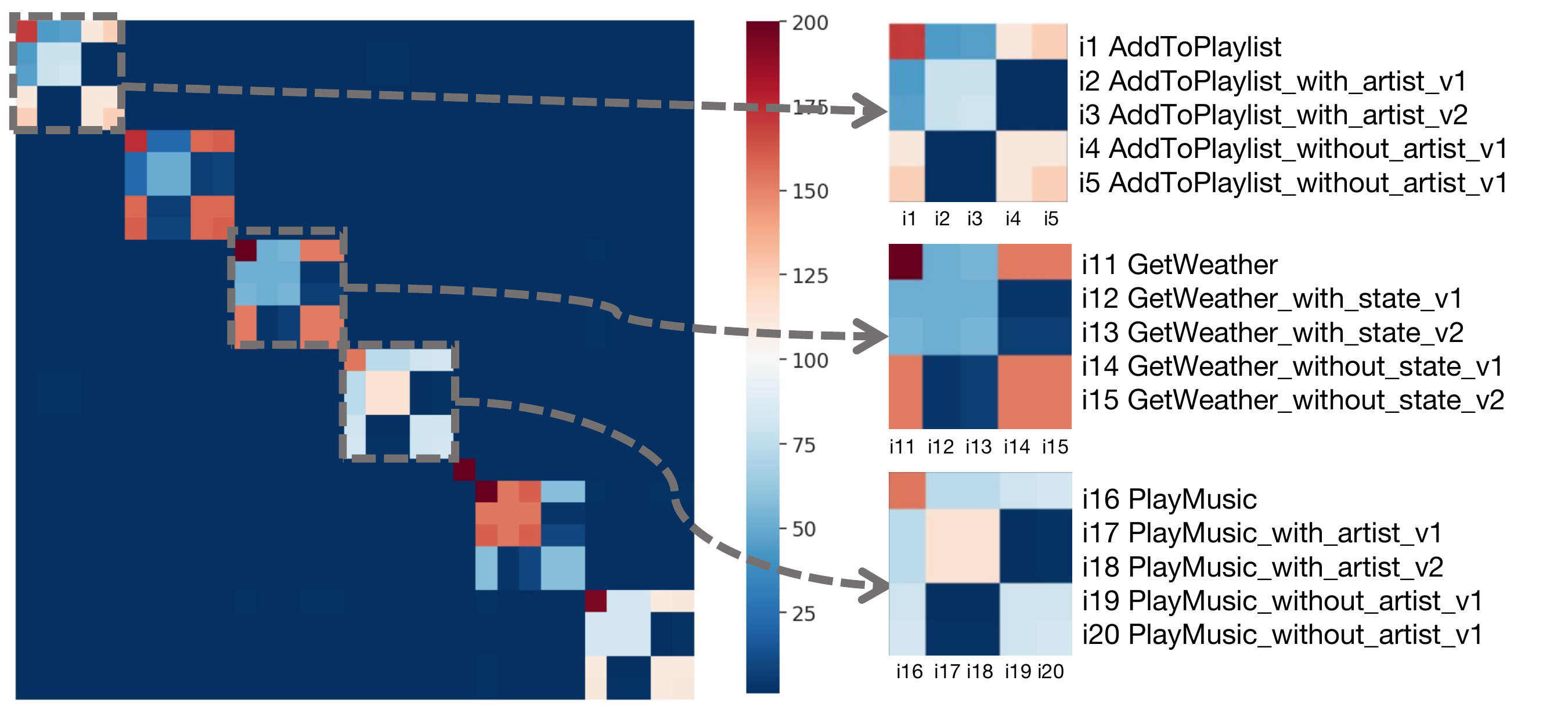}    
    \caption{
    Display of the co-occurrence matrix between labels based on the model output of Multi-Label Focal Loss method for the test set of SNIPS-VCS. Different colors indicate different co-occurrence frequency of labels. The proposed method is able to capture the potential correlation between labels as the model output distinctly corresponds to the relationship between labels, i.e. the frequency of co-occurrence between $i_1^{v_1}$, $i_1^{v_2}$, $i_2^{v_1}$, $i_2^{v_2}$ and $i_1 \& i_2$ is significantly higher than the other labels.
    }
    \label{figure:correlation}
\end{figure*}

Due to the discrepancy between training set and test set for the proposed task, the key point for model success is to capture the potential correlation between related labels, i.e., labels of $i_1^{v_1}$, $i_1^{v_2}$, $i_2^{v_1}$, $i_2^{v_2}$ and $i_1 \& i_2$. Figure~\ref{figure:correlation} displays the co-occurrence matrix between labels based on the model output of Multi-Label Focal Loss method for the test set of SNIPS-VCS. The proposed method is able to capture the potential correlation between labels as the model output distinctly corresponds to the relationship between labels, i.e. the frequency of co-occurrence between $i_1^{v_1}$, $i_1^{v_2}$, $i_2^{v_1}$, $i_2^{v_2}$ and $i_1 \& i_2$ is significantly higher than the other labels. We also visualize the model's prediction on different version labels in the test set of SNIPS-VCS in Appendix~\ref{appendix:visualization}.

\section{Related Work}
\paragraph{Robust NLU}
In the recent years, the topics concerning the NLP robustness and debiasing have attracted board attention. \cite{liu2020empirical,liu2020hyponli,wang2021behind}
For NLU models, \citet{nechaev2021towards} studied data-efficient techniques to make NLU models robust to ASR errors, including data augmentation, adversarial training, and a confidence-aware layer.
\citet{fang2020using} proposed novel phonetic-aware text representations which represent ASR transcriptions at the phoneme level, aiming to
capture pronunciation similarities.
Besides ASR, there are other factors that affect the robustness of the NLU systems.
\citet{liu2020robustness} analyzed different factors affecting the robustness of NLU models including language variety, speech characteristics, and noise perturbation.
\citet{ghaddar2021end} proposed a debiasing framework to slove out-of-distribution (OOD) problem in NLU. 
\citet{zhang2021towards} discussed three robustness problems, namely poor generalization across domains, inherently ambiguous training samples, and unreliable datasets.
To the best of our knowledge, this study is the first to investigate the non-robustness of NLU systems caused by overlapping and conflicting labels resulting from continuous system updates.

\paragraph{Multi-label classification}
Multi-label classification \cite{tsoumakas2006review,zhang2013review,liu2021emerging,wang-etal-2022-hpt} is a well-studied problem that allows each sample assigned multiple labels simultaneously. 
The simplest solution is converting the multi-label problem into multiple independent binary classifications (one for each label) \cite{liu2017deep}. 
But different labels are generally correlated with each other, instead of being independent. 
Some methods are proposed to exploit label correlations in multi-label classification \cite{zhang2010multi,sun2010multi,kong2014large,zhu2017multi}. 
Additionally, there are some studies treating the task as a ranking problem, trying to rank all positive labels higher than other labels for each sample \cite{gong2013deep,kanehira2016multi}. 
All of these works assume that each instance in training data is fully assigned without any missing labels. 
However, the label assignments can be incomplete in many real-world scenarios, especially with a large label set. 

\paragraph{PU Learning}
The label incomplete problem is related to positive and unlabeled (PU) learning \cite{bekker2020learning}. 
PU learning aims to train a classifier from a set of positive samples and an additional set of unlabeled samples.
Many works focus on identifying reliable negative examples from the unlabeled dataset and utilize the estimated labels to improve the classification performances \cite{chaudhari2012learning,ienco2012context,basile2017density,he2018instance}. 
Biased PU learning methods treat the unlabeled samples as negative samples with noise, and use higher penalties on misclassified positive samples to accommodate noise \cite{liu2003building,ke2012building}.
Most studies on PU learning concentrate on binary classification problems which are not sufficient to cover the wide range of real-world applications. 
\citet{xu2017multi} proposed a one-step method that directly enables a multi-class model to be trained using the given multi-class PU data.
Furthermore, there are relatively few studies that explore PU learning for multi-label tasks \cite{sun2010multi,kong2014large,kanehira2016multi,han2018multi}.
\citet{cole2021multi} addressed the hardest multi-label version in which there is only a single positive label available for each sample in training time, and the model needs to predict all proper labels at test time.

\section{Conclusion}
The version conflicts of intents occur frequently due to the semantic overlapping between emerging and existing intents in the industrial dialogue system updates, but are unexplored in the research community. We take a first step to model the version conflict problem as a multi-label classification with positive but unlabeled intents, and propose a dialogue version control (DialogVCS) benchmark with extensive baselines. We find that the overlapping intents can be effectively detected with an automatic workflow.

\section*{Limitations}
In this paper, we focused on the version conflicts of the intents in the NLU model update, without considering dataset noise or skewed intent distribution (extreme long-tail intents). In the real-world applications, other problems would appear in the same time as the version conflicts, thus largely impeding the robustness of NLU models. We call for more realistic, product-driven datasets for more in-depth analyses of the robustness of NLU models.

\section*{Ethics Statement}
The raw data we used to create the dialogue version control datasets (DialogVCS) are all publicly available. We employ automatic data process to simulate the semantic overlapping problem as new intents emerge in the NLU model update, without introducing new user utterances. We guarantee that no user privacy or any other sensitive data being exposed, and no gender/ethnic biases, profanities would appear in the proposed DialogVCS benchmark. The model trained with the benchmark is used to identify the overlapping intents and would not generate any malicious content.

\bibliography{custom}
\bibliographystyle{acl_natbib}

\clearpage
\newpage

\appendix

\section{Appendix}
\renewcommand\thefigure{\Alph{section}\arabic{figure}}    
\renewcommand\thetable{\Alph{section}\arabic{table}}    
\setcounter{table}{0}
\setcounter{figure}{0}

\subsection{Hyper Parameters}
\label{appendix:hyper_parameters}

We list the detailed hyperparameters in Table \ref{table:appendix_hyper_parameters}.
All experiments are run on a NVIDIA-A40. In Table \ref{tab:appendix_model_map}, we list the models used in this paper and their mapping with the hugginface model\_ids. We use a NVIDIA-A40 for 80 hours to get all the reported results.

\begin{table*}
\small
\setlength\tabcolsep{3pt}
    \centering
    \begin{tabular}{lcccc}
    \toprule
    Name & \textbf{ATIS-VCS} & \textbf{SNIPS-VCS} & \textbf{CrossWOZ-VCS} & \textbf{MultiWOZ-VCS}\\
    \midrule
    Learning Rate &{2e-5} &{2e-5} &{2e-5} &{2e-5} \\
    Batch Size &{512} &{512} &{512} &{512}\\
    Max Sequence Length &{32} &{32} &{32} &{32} \\
    Sample Number in Sec.\ref{subsection:method_against_incorrect_negative_label} &{4} & {1} & {1} & {1} \\
    $\mathbf{\beta}$ in Eq.\ref{equation:label_smoothing} & {0.2} &{0.1} &{0.1} &{0.4} \\
    $\mathbf{\gamma}$ in Eq.\ref{equation:focal_loss} & {4} & {4} & {4} & {4} \\
    $\mathbf{\alpha}_{\text{neg}}$ in Eq.\ref{equation:focal_loss} & {0.00001} & {0.00001} & {0.00001} & {0.00001} \\
    $\mathbf{\alpha}_{\text{pos}}$ in Eq.\ref{equation:focal_loss} & {0.99999} & {0.99999} & {0.99999} & {0.99999} \\
    $\mathbf{s}_{0}$ in Eq.\ref{equation:multi-label-cross-entropy-threshold} & {0} & {0} & {0} & {0} \\
    \bottomrule
    \end{tabular}
    \caption{All hyper parameters used in Table \ref{table:main-results}.}
    \label{table:appendix_hyper_parameters}

\end{table*}

\begin{table}[]
    \centering
    \small
    \begin{tabular}{cc}
    \hline
        Model\_name & Hugginface\_ModelID \\
    \hline
        BERT-small (English) & bert-small \\
        BERT-base (English)  & bert-base-uncased \\
        BERT-large (English)  & bert-large-uncased \\
        RoBERTa-base (English) & roberta-base \\
        ALBERT-base (English) & albert-base-v2 \\
        DeBERTa-base (English) & deberta-base \\
        BERT-small (Chinese) & bert-tiny \\
        BERT-base (Chinese)  & bert-base-chinese \\
        BERT-large (Chinese)  & bert-large-chinese \\
        RoBERTa-base (Chinese) & chinese-roberta-wwm-ext \\
        \multirow{2}{*}{ALBERT-base (Chinese)} & albert-base-chinese\\
        & -cluecorpussmall \\
        DeBERTa-base (Chinese) & deberta-base-chinese \\
    \hline
    \end{tabular}
    \caption{The model mapping between model names and hugginface model ids used in this paper.}
    \label{tab:appendix_model_map}
\end{table}

\subsection{Split intent in Proposed Datasets}
\label{appedix:appendix_split}

For single-intent datasets ATIS and SNIPS, we split the intent into two sub-intents by critical entity, which is listed in Table \ref{table:split_intent}. For multi-intent datasets MultiWOZ and CrossWOZ, we split the composite intent into several atomic intents, which is listed in Table \ref{table:split_multiwoz_crosswoz}.
\begin{table}[h]
\begin{subtable}[]{0.5\textwidth}
\small
\centering
\begin{tabular}{p{4cm}l}
\hline
\textbf{Intent} & \textbf{Split Entity} \\ \hline
flight & time \\
abbreviation & fare\_basis\_code \\
aircraft & loc \\
airfare & cost\_relative \\
airline & airline\_code \\
capacity & aircraft\_code \\
city & airline\_name \\
flight\_no & airline\_name \\
flight\_time & depart \\ 
ground\_service & airport\_name \\
\hline
\end{tabular}
\caption{ATIS-VSC}
\label{table:split_ATIS}
\end{subtable}

\begin{subtable}[]{0.5\textwidth}
\small
\centering
\begin{tabular}{p{4cm}l}
\hline
\textbf{Intent} & \textbf{Split Entity} \\ \hline
AddToPlaylist & artist \\
BookRestaurant & restaurant\_name \\
GetWeather & state \\
PlayMusic & artist \\
SearchCreativeWork & object\_type \\
SearchScreeningEvent & object\_type \\ \hline
\end{tabular}
\caption{SNIPS-VSC}
\label{table:split_SNIPS}
\end{subtable}
\caption{Split intent of ATIS (\ref{table:split_ATIS}) and SNIPS (\ref{table:split_SNIPS})}
\label{table:split_intent}
\end{table}

\begin{table}[h]
\small
\begin{subtable}[]{0.5\textwidth}
\centering
\begin{tabular}{p{3.5cm}p{3cm}}
\hline
\textbf{Composite Intent} & \textbf{Atomic Intent} \\
\hline
attraction\&hotel & attraction,hotel \\
attraction\&restaurant & attraction,restaurant \\
attraction\&train & attraction,train \\
hotel\&restaurant & hotel,restaurant \\
hotel\&taxi & hotel,taxi \\
hotel\&train & hotel,train \\
restaurant\&taxi & restaurant,taxi \\
restaurant\&train & restaurant,train \\
\hline
\end{tabular}
\caption{MultiWOZ-VSC}
\label{table:split_multiwoz}
\end{subtable}

\begin{subtable}[]{0.5\textwidth}
\centering
\begin{tabular}{p{3.5cm}p{3cm}}
\hline
\textbf{Composite Intent} & \textbf{Atomic Intent} \\
\hline
General\&Inform & General,Inform \\
General\&Inform\&Request & General,Inform,Request \\
General\&Inform\&Select & General,Inform,Select \\
General\&Request & General,Request \\
Inform\&Request & Inform,Request \\
Inform\&Request\&Select & Inform,Request,Select \\
Inform\&Select & Inform,Select \\
\hline
\end{tabular}
\caption{CrossWOZ-VSC}
\label{table:split_crosswoz}
\end{subtable}
\caption{Split intent of MultiWOZ (\ref{table:split_multiwoz}) and CrossWOZ (\ref{table:split_crosswoz}) }
\label{table:split_multiwoz_crosswoz}
\end{table}

\begin{table}[ht!]
\small
\centering
\begin{tabular}{lcccc}
\hline
\textbf{Dataset} & \textbf{Difficulty} & \textbf{VC-N} & \textbf{MF-N} & \textbf{Total} \\ \hline
ATIS\_1 & Easy 1 & 4 & 1 & 20 \\
ATIS\_2 & Easy 2 & 8 & 2 & 24 \\
ATIS\_4 & Easy 4 & 16 & 4 & 32 \\
ATIS & Normal & 50 & 10 & 66 \\
\hline
SNIPS\_1 & Easy 1 & 4 & 1 & 11 \\
SNIPS\_2 & Easy 2 & 8 & 2 & 15 \\
SNIPS\_4 & Easy 4 & 16 & 4 & 23 \\
SNIPS & Normal & 24 & 6 & 31 \\
\hline
MultiWOZ\_1 & Hard 1 & 4 & 1 & 17 \\
MultiWOZ\_2 & Hard 2 & 6 & 2 & 18 \\
MultiWOZ\_4 & Hard 4 & 10 & 4 & 20 \\
MultiWOZ & Normal & 14  & 8 & 22 \\
\hline
CrossWOZ\_1 & Hard 1 & 4 & 1 & 15 \\
CrossWOZ\_2 & Hard 2 & 6 & 2 & 17 \\
CrossWOZ\_ & Hard 4 & 8 & 4 & 17 \\
CrossWOZ & Normal & 10 & 7 & 17 \\
\hline
\end{tabular}
\caption{The number of version conflict labels (VC-N), merge friction labels (MF-N), and the total labels (Total) of the proposed datasets according to the difficulty levels. The difficulty levels are paired with the ones in Table \ref{table:result_difficulty_control}. ``Easy $k$'' or ``Hard $k$'' means there are $k$ group of version labels.}
\label{table:dataset_label_count_difficulty}
\end{table}

\subsection{Visualization}
\label{appendix:visualization}

\begin{figure*}[h]
    \centering
    \includegraphics[width=1.0\linewidth]{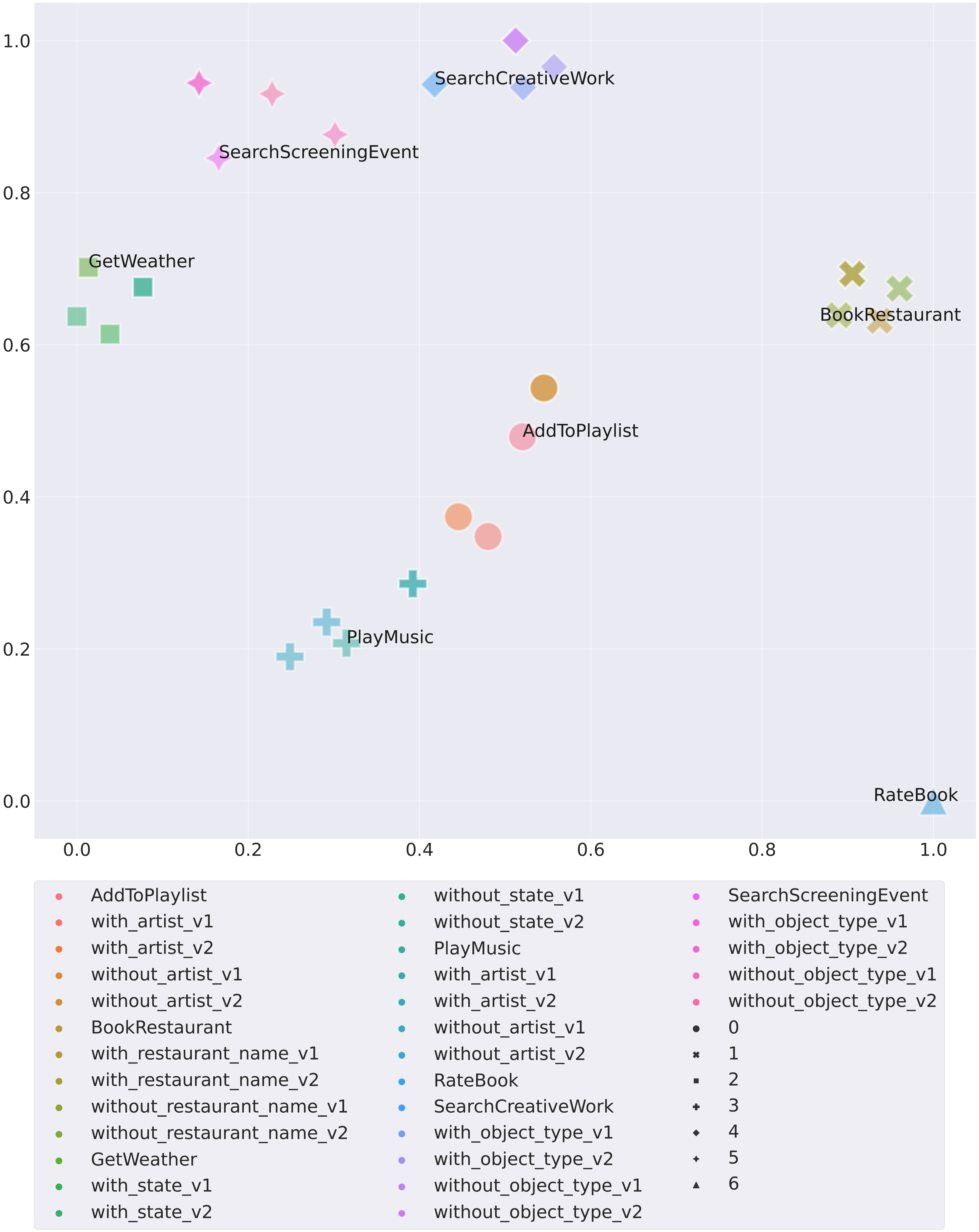}    
    \caption{
    t-SNE dimensionality reduction and DBSCAN clustering for SNIPS. Different colors represent different intents while different shape reperesent differt clusters.
    }
    \label{fig:snips_tsne_dbscan}
\end{figure*}

In Figure \ref{fig:snips_tsne_dbscan}, we visualize the model's ``behavior'' on different version labels in the test set of SNIPS-VCS. Different colors represent different labels, while different shapes represent different clusters. From the figure, we can see that different versions of the same intent family are clustered together. We first use t-SNE \citep{JMLR:v9:vandermaaten08a} to reduce the co-occurrence matrix to two dimensions, then use DBSCAN \citep{10.5555/3001460.3001507} to cluster the labels.

\subsection{Extended Experiment Results}
We list the full experiment scores of the analyses on model scale up, model structure, label smoothing for Label-Smoothing Focal Loss, negative sample number in Table \ref{table:appendix_scale_up}, \ref{table:appendix_model_structure}, \ref{table:appendix_label_smothing}, \ref{table:appendix_negative_sample_number}, respectively.

\begin{table*}
\centering
\small
\setlength\tabcolsep{2.2pt}
    \centering
\begin{tabular}{l|cccc|cccc|cccc|cccc}
\toprule
\multirow{2}{*}{\textbf{Size}}  & 
\multicolumn{4}{c|}{\textbf{ATIS-VCS}}  & 
\multicolumn{4}{c|}{\textbf{SNIPS-VCS}} &
\multicolumn{4}{c|}{\textbf{CrossWOZ-VCS}} &
\multicolumn{4}{c}{\textbf{MultiWOZ-VCS}} \\
  & P & R & F1 & EM & P & R & F1 & EM & P & R & F1 & EM & P & R & F1 & EM \\
\midrule
BERT-small & 66.28 & 94.10 & 77.78 & 47.37 & 85.67 & 96.32 & 90.68 & 76.57 & 93.60 & 97.70 & 95.60 & 90.23 & 82.68 & 90.16 & 86.26 & 81.73 \\
BERT-base & 84.17 & 88.81 & 86.43 & 77.05 & 95.85 & 95.95 & 95.90 & 92.86 & 97.00 & 88.37 & 92.48 & 80.34 & 88.62 & 86.45 & 87.52 & 85.85 \\
BERT-large & 87.14 & 96.46 & 91.57 & 79.34 & 97.32 & 97.58 & 97.45 & 96.71 & 97.35 & 79.72 & 87.66 & 68.69 & 88.60 & 86.11 & 87.34 & 85.85 \\
\bottomrule
\end{tabular}
    \caption{Additional study on different size of BERT including BERT-Small, BERT-Base and BERT-Large. We use Label-Smoothing Focal Loss method to get all the results. Metrics in this table are Precision, Recall, F1-Score and Exact Match Ratio.}
    \label{table:appendix_scale_up}
\end{table*}

\begin{table*}
\centering
\small
\setlength\tabcolsep{2.2pt}
    \centering
\begin{tabular}{l|cccc|cccc|cccc|cccc}
\toprule
\multirow{2}{*}{\textbf{Model}}  & 
\multicolumn{4}{c|}{\textbf{ATIS-VCS}}  & 
\multicolumn{4}{c|}{\textbf{SNIPS-VCS}} &
\multicolumn{4}{c|}{\textbf{CrossWOZ-VCS}} &
\multicolumn{4}{c}{\textbf{MultiWOZ-VCS}} \\
  & P & R & F1 & EM & P & R & F1 & EM & P & R & F1 & EM & P & R & F1 & EM \\
\midrule
BERT-base & 84.17 & 88.81 & 86.43 & 77.05 & 95.85 & 95.95 & 95.90 & 92.86 & 97.00 & 88.37 & 92.48 & 80.34 & 88.62 & 86.45 & 87.52 & 85.85 \\
RoBERTa-base & 87.37 & 95.02 & 91.03 & 79.91 & 96.42 & 96.26 & 96.34 & 95.43 & 94.90 & 90.04 & 92.41 & 79.80 & 85.94 & 87.30 & 86.62 & 83.86 \\
AlBERT-base & 88.10 & 81.43 & 84.64 & 68.95 & 91.69 & 85.42 & 88.45 & 74.71 & 96.83 & 75.08 & 84.58 & 68.69 & 86.35 & 85.50 & 85.92 & 84.27 \\
DeBERTa-base & 90.52 & 92.62 & 91.56 & 85.39 & 96.90 & 85.58 & 90.89 & 75.14 & 96.40 & 95.46 & 95.93 & 88.08 & 88.99 & 84.3 & 86.61 & 80.75 \\
\bottomrule
\end{tabular}
    \caption{ Results of four models including BERT, RoBERTa, AlBERT and DeBERTa. We Label-Smoothing Focal Loss method to get all the reported results. Metrics in this table are Precision, Recall, F1-Score and Exact Match Ratio.}
    \label{table:appendix_model_structure}
\end{table*}

\begin{table*}
\centering
\small
\setlength\tabcolsep{2.2pt}
    \centering
\begin{tabular}{l|cccc|cccc|cccc|cccc}
\toprule
\multirow{2}{*}{\textbf{LSR}}  & 
\multicolumn{4}{c|}{\textbf{ATIS-VCS}}  & 
\multicolumn{4}{c|}{\textbf{SNIPS-VCS}} &
\multicolumn{4}{c|}{\textbf{CrossWOZ-VCS}} &
\multicolumn{4}{c}{\textbf{MultiWOZ-VCS}} \\
  & P & R & F1 & EM & P & R & F1 & EM & P & R & F1 & EM & P & R & F1 & EM \\
\midrule
0.1 & 65.99 & 94.37 & 77.67 & 47.72 & 95.85 & 95.95 & 95.90 & 92.86 & 97.00 & 88.37 & 92.48 & 80.34 & 85.26 & 88.53 & 86.86 & 83.75 \\
0.2 & 84.17 & 88.81 & 86.43 & 77.05 & 96.63 & 93.53 & 95.05 & 89.00 & 95.53 & 70.14 & 80.89 & 40.66 & 86.88 & 87.16 & 87.02 & 84.58 \\
0.4 & 91.59 & 79.53 & 85.13 & 73.63 & 97.41 & 81.21 & 88.58 & 65.86 & 95.34 & 69.79 & 80.59 & 40.23 & 88.62 & 86.45 & 87.52 & 85.85 \\
\bottomrule
\end{tabular}
    \caption{ Results of different label smoothing rate used in Label-Smothing Focal Loss including 0.1, 0.2, and 0.4. We use Label-Smoothing Focal Loss method to get all the reported results. Metrics in this table are Precision, Recall, F1-Score, and Exact Match Ratio.}
    \label{table:appendix_label_smothing}
\end{table*}

\begin{table*}
\centering
\small
\setlength\tabcolsep{2.2pt}
    \centering
\begin{tabular}{l|cccc|cccc|cccc|cccc}
\toprule
\multirow{2}{*}{\textbf{NSN}}  & 
\multicolumn{4}{c|}{\textbf{ATIS-VCS}}  & 
\multicolumn{4}{c|}{\textbf{SNIPS-VCS}} &
\multicolumn{4}{c|}{\textbf{CrossWOZ-VCS}} &
\multicolumn{4}{c}{\textbf{MultiWOZ-VCS}} \\
  & P & R & F1 & EM & P & R & F1 & EM & P & R & F1 & EM & P & R & F1 & EM \\
\midrule
1 & 50.46 & 96.84 & 66.35 & 55.59 & 94.30 & 93.16 & 93.73 & 85.14 & 97.97 & 49.24 & 65.54 & 42.97 & 86.19 & 87.06 & 86.62 & 82.79 \\
2 & 69.95 & 92.20 & 79.55 & 67.92 & 95.97 & 87.74 & 91.67 & 74.71 & 97.88 & 42.01 & 58.78 & 35.61 & 88.28 & 81.15 & 84.57 & 74.50 \\
4 & 87.40 & 86.87 & 87.14 & 76.37 & 97.00 & 71.58 & 82.37 & 41.14 & 97.81 & 36.25 & 52.90 & 28.41 & 90.15 & 68.10 & 77.59 & 50.65 \\
8 & 93.00 & 77.36 & 84.46 & 71.23 & 96.96 & 63.84 & 76.99 & 32.57 & 97.67 & 32.45 & 48.72 & 22.42 & 91.34 & 60.23 & 72.60 & 35.71 \\
\bottomrule
\end{tabular}
    \caption{ Results of five Negative Sample number including 1, 2, 4 and 8. We use NS method to get all the reported results. Metrics in this table are Precision, Recall, F1-Score, and Exact Match Ratio.}
    \label{table:appendix_negative_sample_number}
\end{table*}

\subsection{Difficulty Control}
\label{appendix:difficulty_control}

We introduce version conflict and merge friction to every possible label, but in practice, we may not see version labels in such a high proportion. To better simulate the actual scenario and also have better control over the difficulty of the datasets, we limit the number of version labels to 1, 2, and 4. For ATIS-VCS and SNIPS-VCS, more version labels would be more difficult, since intent splitting creates sub-intents that need to check both the original intent and the critical entity. For example, checking the sub-intent ``$Flight\_with\_time$'' requires more computation than full-intent ``$Flight$''.However, for MultiWOZ-VCS and CrossWOZ-VCS, more version labels would not be more difficult, because composite-intent splitting creates atomic intents that are easier to check. Fore example, checking the composite intent ``$Hotel\&Taxi$'' requires more computation them simply checking atomic intent ``$Hotel$'' or ``$Taxi$''. The statistics is shown in Table \ref{table:dataset_label_count_difficulty}. 

\begin{table*}
\small
\begin{minipage}{\textwidth}
\centering
\makeatletter\def\@captype{table}
\begin{tabular}{l|cccc|cccc}
\toprule
\multirow{2}{*}{\textbf{Difficulty}}  & 
\multicolumn{4}{c|}{\textbf{ATIS-VCS}}  & 
\multicolumn{4}{c}{\textbf{SNIPS-VCS}} \\
  & P & R & F1 & EM & P & R & F1 & EM \\
\midrule
Easy 1 & 93.90 & 98.55 & 96.17 & 91.44 & 98.34 & 98.67 & 98.50 & 98.43 \\
Easy 2 & 94.47 & 98.32 & 96.46 & 92.35 & 96.42 & 95.45 & 95.93 & 94.71 \\
Easy 3 & 88.15 & 98.75 & 93.15 & 82.99 & 96.98 & 96.47 & 96.72 & 95.29 \\
Normal & 84.17 & 88.81 & 86.43 & 77.05 & 95.85 & 95.95 & 95.90 & 92.86 \\
\bottomrule
\end{tabular}
\label{sample-table}
\makeatletter\def\@captype{table}
\begin{tabular}{l|cccc|cccc}
\toprule
\multirow{2}{*}{\textbf{Difficulty}}  & 
\multicolumn{4}{c|}{\textbf{CrossWOZ-VCS}}  & 
\multicolumn{4}{c}{\textbf{MultiWOZ-VCS}} \\
  & P & R & F1 & EM & P & R & F1 & EM \\
\midrule
Hard 1 & 94.50 & 63.65 & 76.07 & 44.10 & 81.28 & 88.98 & 84.96 & 82.20 \\
Hard 2 & 95.53 & 70.14 & 80.89 & 40.66 & 82.93 & 88.04 & 85.41 & 83.36 \\
Hard 3 & 95.34 & 69.79 & 80.59 & 40.23 & 84.68 & 87.96 & 86.29 & 83.51\\
Normal & 97.00 & 88.37 & 92.48 & 80.34 & 88.62 & 86.45 & 87.52 & 85.85 \\
\bottomrule
\end{tabular}
\end{minipage}
\caption{Results of 3 difficulty including 1, 2 and 4 in the four datasets: ATIS, SNIPS, CrossWOZ and MultiWOZ. Metrics in this table are F1-Score, Exact Match Ratio and Zero One Loss. 1 is the easiest and 4 is hardest.}
\label{table:appendix_result_difficulty_control}
\end{table*}


\end{document}